\documentclass{article}

\PassOptionsToPackage{numbers, compress}{natbib}
\usepackage[preprint]{neurips_2024}

\usepackage[utf8]{inputenc} %
\usepackage[T1]{fontenc}    %
\usepackage{hyperref}       %
\usepackage{url}            %
\usepackage{booktabs}       %
\usepackage{amsfonts}       %
\usepackage{nicefrac}       %
\usepackage{microtype}      %
\usepackage{xcolor}         %

\usepackage{caption}

\usepackage{amsmath}
\usepackage{amssymb}
\usepackage{mathtools}
\usepackage{amsthm}

\usepackage{multirow} 
\usepackage{booktabs}
\usepackage{wrapfig}
\usepackage{bm}
\usepackage{subcaption}
\usepackage{xspace}
\usepackage{colortbl}
\usepackage{float}
\usepackage{cleveref}

\hypersetup{
    colorlinks,
    linkcolor={red!70},
    citecolor={green!50!black},
    urlcolor={blue!80!black}
}
\usepackage[misc]{ifsym} %

\makeatletter
\DeclareRobustCommand\onedot{\futurelet\@let@token\@onedot}
\def\@onedot{\ifx\@let@token.\else.\null\fi\xspace}

\def\eg{\emph{e.g}\onedot} 
\def\ie{\emph{i.e}\onedot} 
 
 \def\vs{\emph{vs}\onedot}

\makeatother
\newcommand{\name}{ViG}

\crefformat{section}{\S#2#1#3} %
\crefformat{subsection}{\S#2#1#3}
\crefformat{subsubsection}{\S#2#1#3}

\newcommand{\rblue}{\rowcolor{blue!10}}
\newcommand{\boldparagraph}[1]{\vspace{-0.1cm}\noindent{\bf #1}}
\title{ViG: Linear-complexity Visual Sequence Learning with Gated Linear Attention}

\author{%
  Bencheng Liao$^{1, 2, \diamond}$ 
  \ \ \ \ 
  Xinggang Wang$^{2~\textrm{\Letter}}$
  \ \ \ \ 
  Lianghui Zhu$^{2}$ 
  \ \ \ \
  Qian Zhang$^{3}$ 
  \ \ \ \
  \vspace{.5em}
  Chang Huang$^{3}$
  \\
  $^{1}$ Institute of Artificial Intelligence, Huazhong University of Science \& Technology
  \\
  $^{2}$ School of EIC, Huazhong University of Science \& Technology
  \\
  $^{3}$ Horizon Robotics
  \\
  \url{https://github.com/hustvl/ViG}
}

\begin{document}

\maketitle
\let\thefootnote\relax\footnotetext{$^\diamond$ Intern of Horizon Robotics when doing this work; $^\boxtimes$ Corresponding author: \texttt{xgwang@hust.edu.cn}}

\begin{abstract}

Recently, linear complexity sequence modeling networks have achieved modeling capabilities similar to Vision Transformers on a variety of computer vision tasks, while using fewer FLOPs and less memory.
However, their advantage in terms of actual runtime speed is not significant.
To address this issue, we introduce Gated Linear Attention (GLA) for vision, leveraging its superior hardware-awareness and efficiency. We propose direction-wise gating to capture  1D global context through bidirectional modeling and a 2D gating locality injection to adaptively inject 2D local details into 1D global context. Our hardware-aware implementation further merges forward and backward scanning into a single kernel, enhancing parallelism and reducing memory cost and latency.
The proposed model, \name{}, offers a favorable trade-off in accuracy, parameters, and FLOPs on ImageNet and downstream tasks, outperforming popular Transformer and CNN-based models. 
Notably, \name{}-S matches DeiT-B's accuracy while using only 27\% of the parameters and 20\% of the FLOPs, running 2$\times$ faster on $224\times224$ images. At $1024\times1024$ resolution, \name{}-T uses 5.2$\times$ fewer FLOPs, saves 90\% GPU memory, runs 4.8$\times$ faster, and achieves 20.7\% higher top-1 accuracy than DeiT-T. These results position \name{} as an efficient and scalable solution for visual representation learning.
\end{abstract}

\section{Introduction}

Vision Transformer (ViT)~\cite{vit} has revolutionized computer vision by introducing an advanced sequence modeling layer Transformer~\cite{vaswani2017attention} from natural language processing (NLP) to perform visual representation learning. It has proven highly successful across various vision tasks~\cite{swin,fang2023eva,fang2023eva02,sun2023eva,radford2021learning,liu2024visual,li2022exploring,xie2021segformer,fang2021you,li2022blip,peebles2023scalable,yu2022metaformer,fang2023unleashing}, serving as a versatile backbone. However, the quadratic complexity inherent in the Transformer's softmax attention presents significant challenges for its applications on high-resolution images. Numerous efforts~\cite{swin,dong2022cswin,yang2021focal} have sought to address this limitation by drawing inspiration from the success of convolutional networks, such as constraining attention computations within local windows. While this approach achieves linear complexity similar to conventional CNNs, it falls short in capturing the global context. This raises a critical question: can we design a fundamental block that combines the best of both worlds—Transformers and CNNs—offering global receptive field and linear complexity?

The recent development of linear-time sequence modeling methods~\cite{mamba,rwkv,gla,qin2024hierarchically,retnet,linearattn} from NLP provides a promising solution to the question. These methods operate similarly to RNNs by compressing all historical inputs into a fixed-size state and then attending to the current input based on this compressed state, unlike Transformers~\cite{vaswani2017attention} which attend to all historical states. To further address the wall-time efficiency limitations of explicit recurrent forms, Mamba~\cite{mamba}, RWKV~\cite{rwkv}, and GLA~\cite{gla} introduce hardware-aware implementations, demonstrating superior accuracy and efficiency compared to highly-optimized Transformers.
Mamba, in particular, has been widely adopted and adapted for vision tasks. While these methods achieve superior classification accuracy at a resolution of 224 and exhibit impressive efficiency in terms of FLOPs and memory at high resolutions (\eg, 1248 resolution in Vision Mamba), their wall-time efficiency at lower and more common resolutions is often comparable to or even less than that of ViT/DeiT~\cite{vit,deit}.

This limitation inspires us to explore the application of the superior hardware-efficient GLA~\cite{gla,fla} to push the efficiency and accuracy envelope of linear-complexity visual sequence learning, making it more practical and competitive with the well-established and highly-optimized Transformers and CNNs. Different from Mamba based on SISO (single-input-single-output) state space model~\cite{gu2021efficiently}, GLA originates from linear attention~\cite{linearattn}, which has a simple and hardware-friendly matrix-multiply form by approximating softmax in standard attention with a linear kernel. To further enhance the expressiveness, GLA introduces a novel data-dependent gating mechanism to adaptively control the forget rate of compressed state.  

However, the vanilla GLA model is designed for unidirectional modeling, which has temporal dependence. This makes its ability for global perception in NLP not truly global when applied to vision tasks. 
Despite some follow-up works~\cite{zigma,yang2024plainmamba,localmamba} on Mamba in vision that explore additional scanning directions beyond bidirectional modeling of 1D visual sequences to approximate vanilla attention's ability to interact with visual tokens in any direction and position, these approaches suffer from significant memory inefficiencies due to frequent, non-sequential memory access patterns, making them less practical in terms of wall-time efficiency. We adhere to bidirectional modeling for its simplicity and memory-friendly access pattern.
To further harness the inherent directional sensitivity in vision data (\ie, information from different visual directions varies significantly in importance), we propose a bidirectional GLA (BiGLA) layer by designing a direction-wise gating mechanism to adaptively select the global context from different directions. This design shares most parameters between the forward and backward directions. Though the proposed directional design achieves global context along the 1D visual sequence, it still fails to capture the 2D nature of visual data. To address this, we propose a 2D gating locality injection to adaptively compress 2D local information extracted by convolution into the 1D global context extracted by the sequence modeling BiGLA layer. Moreover, the proposed parameter-efficient bidirectional design allows us to merge bidirectional scanning into a single kernel, enhancing the hardware-awareness of the implementation and reducing memory cost and latency introduced by the extra direction.

\begin{figure}[t!]

    \centering
     \begin{subfigure}[b]{0.47\textwidth}
         \centering
         \includegraphics[width=\textwidth]{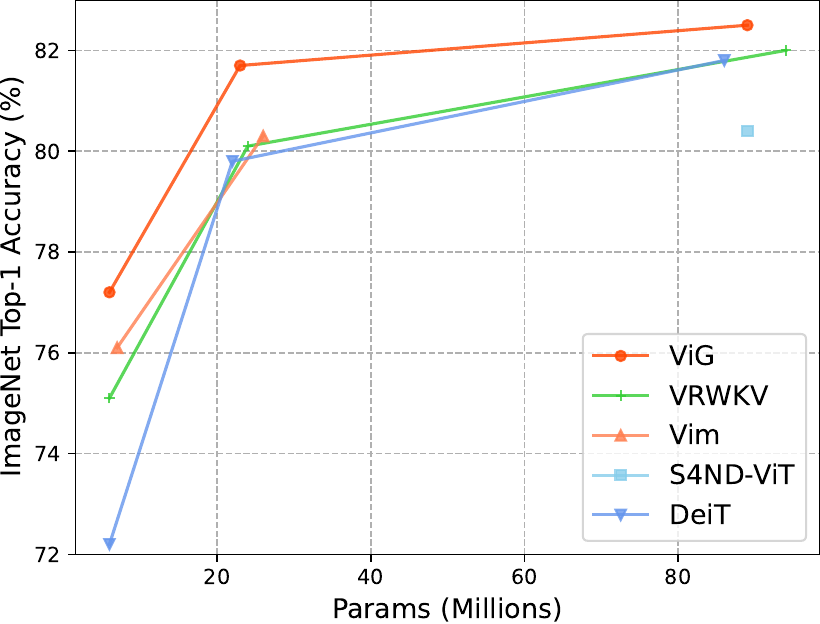}
         \vspace{-0.5cm}
         \caption{Comparison with non-hierarchical architectures}
     \end{subfigure}
     \hspace{-0.1cm}
     \begin{subfigure}[b]{0.48\textwidth}
         \centering
         \includegraphics[width=\textwidth]{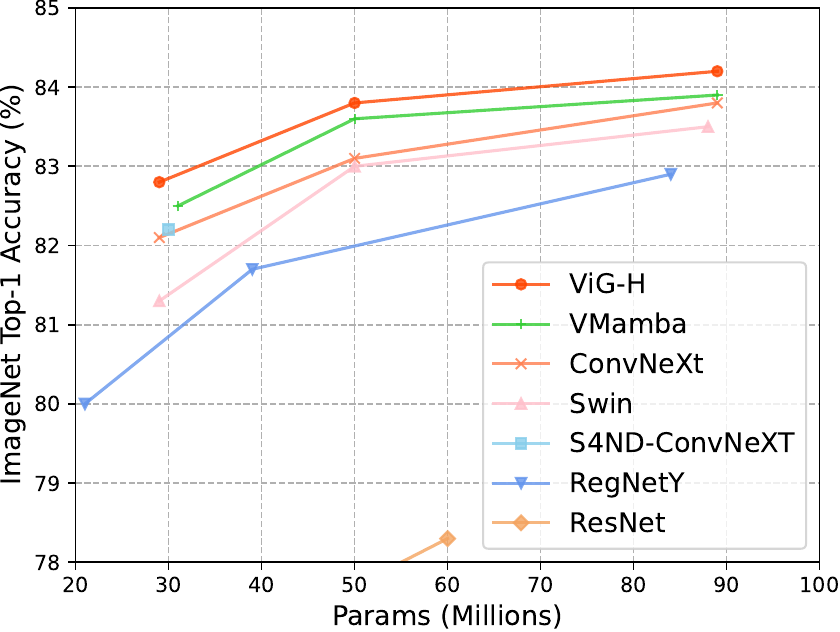}
         \vspace{-0.5cm}
         \caption{Comparison with hierarchical architectures}
        \end{subfigure}
        \vspace{-0.2cm}
        \caption{Performance comparisons of (a) non-hierarchical architectures~\cite{vrwkv,deit,vim,nguyen2022s4nd} and (b) hierarchical architectures~\cite{vmamba,convnext,swin,nguyen2022s4nd,regnet,resnet} on ImageNet-1K. Our proposed non-hierarchical ViG and hierarchical ViG-H demonstrate superior performance compared to the popular models in terms of parameters and accuracy. Particularly, the proposed basic ViG block achieves global receptive field with linear complexity, while the CNN~\cite{resnet,regnet,convnext}, vanilla softmax attention~\cite{deit} and window-attention-based~\cite{swin} blocks cannot.} %
        \label{fig:acc_comp} 
\end{figure}

The main contributions of this paper can be summarized as follows:
\begin{itemize}
    \item We present \name{}, a generic vision backbone network that combines the linear complexity of Gated Linear Attention (GLA) with the hardware-awareness needed for efficient visual sequence learning. \name{} addresses the limitations of previous Transformer-based and CNN-based methods, combining the best of both worlds to offer an efficient and scalable solution.
    \item To better adapt to vision tasks, we propose three key designs with minimal overhead: a bidirectional gated linear attention mechanism to capture the global 1D context of visual sequences, a direction-wise gating mechanism to adaptively select global context from different directions, and a 2D gating locality injection to integrate 2D local information into the 1D global context. We further provide a hardware-aware implementation that merges forward and backward scanning into a single kernel, enhancing parallelism and reducing memory cost and latency.
    \item Our models achieve superior performance in terms of accuracy and parameters compared to state-of-the-art non-hierarchical and hierarchical models on ImageNet~\cite{imagenet}, as shown in Fig.~\ref{fig:acc_comp}. For downstream dense prediction tasks~\cite{ade20k,coco}, \name{} outperforms ViT~\cite{vit,deit} and VRWKV~\cite{vrwkv} with lower computational costs across different model sizes. The wall-time efficiency of \name{} outperforms the counter-part linear-complexity visual sequence learning methods~\cite{vim,vrwkv,vmamba} and matches the well-established and highly-optimized ConvNeXt~\cite{convnext} and SwinTransformer~\cite{swin}.
\end{itemize}

\section{Related Work}
\label{sec:related}
ViT~\cite{vit} demonstrated that visual representation learning can be performed in a sequence manner by introducing the Transformer~\cite{vaswani2017attention} from NLP. Many follow-up works~\cite{wu2021cvt,wang2022pvt,yang2021focal,dai2021coatnet,d2021convit,fang2022msg,guo2022cmt,dong2022cswin,tu2022maxvit,zhang2022topformer,liu2022swin,yuan2021tokens,chu2021twins,zhang2022hivit,li2023uniformer} focus on improving ViT's efficiency and performance without altering the softmax attention. Recently, another line of works~\cite{hochreiter1997long,qin2024hierarchically,linearattn,choromanski2020rethinking,arora2024simple,qin2024hgrn2,dai2019transformer,de2024griffin,munkhdalai2024leave,ma2024megalodon} have shown that quadratic softmax attention can be replaced by advanced RNN-like, linear-time sequence modeling methods.
Vision Mamba~\cite{vim} builds upon the linear-time sequence modeling Mamba block~\cite{mamba} by introducing an additional backward SSM layer. VMamba~\cite{vmamba} introduces criss-cross scanning to Mamba and builds a hierarchical architecture. LocalMamba~\cite{localmamba} optimizes scanning directions to exploit local priors for vision. Zigma~\cite{zigma} and PlainMamba~\cite{yang2024plainmamba} introduce multiple scanning directions in a zigzag manner. Many works~\cite{ma2024u,li2024videomamba,chen2024video,liang2024pointmamba,xing2024segmamba,zhao2024rs,zhang2024motion,patro2024simba,shen2024gamba,yang2024vivim,he2024pan,fei2024scalable,chen2024changemamba} have explored Mamba's effectiveness in various vision tasks. In contrast, VisionRWKV~\cite{vrwkv} forgoes the Mamba block, adapting the linear complexity RWKV block from NLP for use in vision.
Extended related works are provided in Appendix~\ref{app_sec:related}.

\section{Preliminary}
In this section, we introduce the evolution from standard softmax attention to advanced gated linear attention (GLA), omitting the multi-head mechanism for simplicity.

\boldparagraph{Softmax Attention.} Softmax attention has been used as a standard block in Transformer~\cite{vaswani2017attention} since it has the strong capability to dynamically model the relationship accross long sequence and enjoys great parallelism for training. Given an input sequence $\mathbf{X} \in \mathbb{R}^{T \times d}$, the softmax attention used in autoregressive Transformer can be defined as:
\begin{align}
\begin{split}
\mathbf{Q},\mathbf{K},\mathbf{V}=\mathbf{X}\bm{W}_{Q},\mathbf{X}\bm{W}_{K},\mathbf{X}\bm{W}_{V},\\
\mathbf{O}=\mathrm{softmax}\big((\mathbf{Q}\mathbf{K}^{\top})\odot\mathbf{M}\big)\mathbf{V},
\end{split}
\end{align}
where $\bm{W}_{Q} \in \mathbb{R}^{d \times d_q}$, $\bm{W}_{K} \in \mathbb{R}^{d \times d_k}$ and $ \bm{W}_{V} \in \mathbb{R}^{d \times d_v}$ are trainable projection matrices, $\mathbf{M} \in \{-\infty,1\}^{T \times T}$ is the causal mask for preventing interaction with future tokens, $\mathbf{O} \in \mathbb{R}^{T \times d_v}$ is the output. The above parallel form can also be written in the following recurrent form to compute the single output $\mathbf{o}_t$:
\begin{align}
\begin{split}
\bm{q}_{t},\bm{k}_{t},\bm{v}_{t}=\bm{x}_{t}\bm{W}_{Q},\bm{x}_{t}\bm{W}_{K},\bm{x}_{t}\bm{W}_{V},\\
\bm{o}_{t}=\frac{\sum_{i=1}^{t}\exp(\boldsymbol{q}_{t}\boldsymbol{k}_{i}^{\top})\boldsymbol{v}_{i}}{\sum_{i=1}^{t}\exp(\boldsymbol{q}_{t}\boldsymbol{k}_{i}^{\top})},
\end{split}
\end{align}
where the output $\mathbf{o}_t$ is computed by attending the projected query $\bm{q}_t$ of $\bm{x}_t$ to the sets of projected keys $\{\bm{k}_1, \dots, \bm{k}_t\}$ and values $\{\bm{v}_1, \dots, \bm{v}_t\}$.

\boldparagraph{Linear Attention.} Linear attention~\cite{linearattn} replaces $\exp(\boldsymbol{q}_{t}\boldsymbol{k}_{i}^{\top})$ in standard softmax attention with feature map dot-products $\phi(\boldsymbol{q}_{t})\phi(\boldsymbol{k}_{i})^{\top}$. Recently, \cite{qin2022devil} has empirically found that linear feature map works well by setting $\phi$ to be the identity and removing the normalizer. This simplifies the computation of $\boldsymbol{o}_t$ as:
\begin{align}
\begin{split}
\bm{o}_t = \boldsymbol{q}_t \sum_{i=1}^{t} \boldsymbol{k}_i^{\top} \boldsymbol{v}_i.
\end{split}
\end{align}
Letting hidden state $\mathbf{S}_{t}=\sum_{i=1}^{t}\boldsymbol{k}_{i}^{\top}\boldsymbol{v}_{i} \in \mathbb{R}^{d_k \times d_v}$, which is fixed-size and compresses the historical information, we can rewrite above computation as an RNN:
\begin{align}
\begin{split}
\mathbf{S}_{t}=\mathbf{S}_{t-1}+\boldsymbol{k}_{t}^{\top}\boldsymbol{v}_{t},\quad\boldsymbol{o}_{t}=\boldsymbol{q}_{t}\mathbf{S}_{t}.
\end{split}
\end{align}

\boldparagraph{Gated Linear Attention.} \cite{gla} propose adding a data-dependent gating mechanism in linear attention to enhance expressiveness. The gated linear attention can be defined as:
\begin{align}
\begin{split}
    \bm{\alpha}_t &= \mathrm{sigmoid}((\bm{x}_t \bm{W}^1_\alpha \bm{W}^2_\alpha + \bm{b}_\alpha))^{\frac{1}{\tau}} \in \mathbb{R}^{1 \times d_k},\\
    \mathbf{G}_t& = \bm{\alpha}_t^{\top} \mathbf{1} \in \mathbb (0,1)^{d_k \times d_v},\\
    \mathbf{S}_t &= \mathbf{G}_t \odot \mathbf{S}_{t-1} + \bm{k}_t^{\top} \bm{v}_t \in \mathbb{R}^{d_k \times d_v}, \\
    \boldsymbol{o}_{t}&=\boldsymbol{q}_{t}\mathbf{S}_{t},
\end{split}
 \end{align}
where $\bm{\alpha}_t$ is obtained from applying a low-rank linear layer on $\bm{x}_t$ followed by sigmoid activation, $\bm{W}^1_\alpha \in \mathbb{R}^{d \times 16}$, $\bm{W}^2_\alpha \in \mathbb{R}^{16 \times d_k}$, $\bm{b}_\alpha \in \mathbb{R}^{1 \times d_k}$ are trainable matrices and $\tau=16 $ is a temperature term to encourage the model to have a slower forgetting rate, $\mathbf{G}_t$ is matrix-form forget gate, expanded by outer-producting with matrice $\mathbf{1}$.

\begin{figure}[t!]
  \begin{center}
      \includegraphics[width=.98\textwidth]{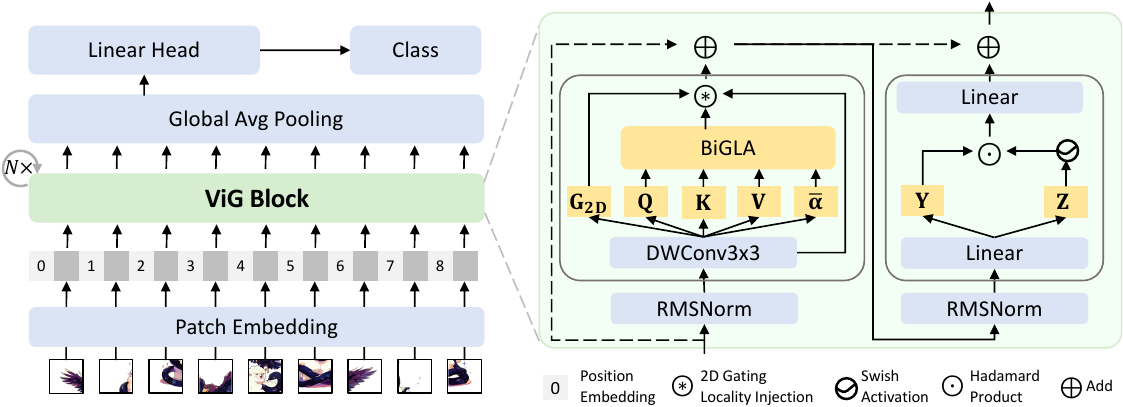}
      \vspace{-0.2cm}
      \caption{The overall architecture of \name{}. We follow ViT~\cite{vit} to build architecture by first transforming the input image into a sequence of patches and then feeding it into $N$ basic \name{} blocks. The proposed \name{} block consists of RMSNorm~\cite{zhang2019root}, the proposed linear complexity spatial mixing layer, and SwiGLU Feed Forward Network~\cite{shazeer2020glu}.}
      \label{fig:framework}
  \end{center}
  \vspace{-0.6cm}
\end{figure}

\section{Method}
\subsection{Overall Architecture}

The overall architecture of our model is depicted in Fig.~\ref{fig:framework}. We first transform the $H \times W \times 3$ image into $T=\frac{H \times W}{p^2}$ patch tokens with $d$ dimensions, where $p$ is the patch size. Before feeding the patch tokens into the stack of \name{} blocks, we add learnable position embeddings. 
The output tokens of the last block are fed into a global average pooling layer followed by a linear classifier.

\subsection{\name{} Block}
\name{} block, serving as a basic block, consists of: 1) a long-term BiGLA layer that can exploit the 1D-global context of the image in a linear-complexity manner; 2) a short-term depth-wise convolution layer that can capture the 2D-local details of the image; 3) a gating mechanism that can adaptively combine the global and local information; 4) a SwiGLU FFN layer as in \cite{touvron2023llama,retnet,gla} for channel mixing.

\subsubsection{Global Bidirectional Gated Linear Attention}
\label{sec:bigla}

In this work, we adopt bidirectional modeling for its inherent simplicity and memory-friendly access pattern. The crux of this design is the direction-wise gating mechanism, particularly the forget gate $\mathbf{G}_t$, which meticulously controls the flow of information. This is crucial, especially for tokens located at the boundaries of an object, where information from different directions varies significantly in importance. To harness this directional sensitivity effectively, we introduce the Bidirectional Gated Linear Attention (BiGLA) layer, as shown in Fig.~\ref{fig:bigla}. This layer is parameter-efficient by sharing all parameters except for the forget gate, which is tailored to each direction:

\begin{minipage}[c]{0.6\textwidth} 
  \vspace*{-0.4cm}
  \begin{align}
    \begin{split}
        \overline{\bm{\alpha}}_t &= \mathrm{sigmoid}((\bm{x}_t \bm{W}^1_\alpha \overline{\bm{W}}^2_\alpha + \overline{\bm{b}}_\alpha))^{\frac{1}{\tau}} \in \mathbb{R}^{1 \times 2d_k}, \\
        \overrightarrow{\bm{\alpha}}_t, \overleftarrow{\bm{\alpha}}_t &= \mathrm{split}(\overline{\bm{\alpha}}_t) \in \mathbb{R}^{1 \times d_k},\\
        \overrightarrow{\mathbf{G}}_t& = \overrightarrow{\bm{\alpha}}_t^{\top} \mathbf{1} \in \mathbb (0,1)^{d_k \times d_v}, \\
        \overleftarrow{\mathbf{G}}_t &= \overleftarrow{\bm{\alpha}}_t^{\top} \mathbf{1} \in \mathbb (0,1)^{d_k \times d_v},\\
        \overrightarrow{\mathbf{S}}_t &= \overrightarrow{\mathbf{G}}_t \odot \overrightarrow{\mathbf{S}}_{t-1} + \bm{k}_t^{\top} \bm{v}_t \in \mathbb{R}^{d_k \times d_v}, \\
        \overleftarrow{\mathbf{S}}_t &= \overleftarrow{\mathbf{G}}_t \odot \overleftarrow{\mathbf{S}}_{t+1} + \bm{k}_t^{\top} \bm{v}_t \in \mathbb{R}^{d_k \times d_v},\\
        \overrightarrow{\boldsymbol{o}}_{t}&=\boldsymbol{q}_{t}\overrightarrow{\mathbf{S}}_{t}, \\
        \overleftarrow{\boldsymbol{o}}_{t}&=\boldsymbol{q}_{t}\overleftarrow{\mathbf{S}}_{t}, \\
        \bm{o}_t &= (\overrightarrow{\boldsymbol{o}}_{t} + \overleftarrow{\boldsymbol{o}}_{t})/2, \label{eq:bigla}
    \end{split}
    \end{align}
    \vspace*{-0.5cm}
\end{minipage}
\hfill 
\begin{minipage}[c]{0.35\textwidth} 
  \centering
  \vspace*{-0.8cm}
  \begin{figure}[H]
      \includegraphics[width=1.\textwidth]{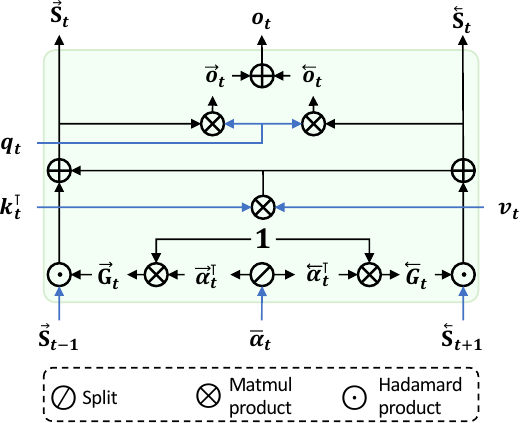} 
      \vspace*{-0.5cm}
      \caption{Illustration of BiGLA.}
      \label{fig:bigla}
  \end{figure}
  \vspace*{-0.8cm}
\end{minipage}

where $\overline{\square}$ indicates bidirectional modification, $\overrightarrow{\square}$ and $\overleftarrow{\square}$ indicate forward and backward direction respectively, $\overline{\bm{W}}^2_\alpha \in \mathbb{R}^{16 \times 2d_k}$ and $\overline{\bm{b}}_\alpha \in \mathbb{R}^{1 \times 2d_k}$.
The proposed BiGLA layer compresses the historical information of forward and backward directions into fixed-size hidden states $\overrightarrow{\mathbf{S}}_t$ and $\overleftarrow{\mathbf{S}}_t$, and attends $\bm{q}_t$ with the hidden states to obtain long-term global context in a linear-complexity manner. 

The proposed design only introduces extra $17d_k$ parameters to render vanilla causal GLA layer into BiGLA layer for visual representation learning, which is minor compared to the total $4d^2$ parameters (roughly) by setting $d_q,d_k=\frac{d}{2}$ and $d_v = d$. The designed BiGLA layer has nearly the same number of parameters as the standard softmax attention while using much fewer FLOPs 
($\Omega_{\text{BiGLA}}=5Td^2+32Td$ \vs $\Omega_{\text{SoftmaxAttn}}=4Td^2+2T^2d$).

\subsubsection{2D Gating Locality Injection} 
\label{sec:2dgating}

Though the hidden states $\overrightarrow{\mathbf{S}}_t$ and $\overleftarrow{\mathbf{S}}_t$ in the BiGLA layer, compressed along the 1D visual sequences, can capture the long-term global context of the image, they may find it difficult in capturing the local details of 2D images. To address this issue, we inject 2D locality by introducing a short-term local convolution layer.
In our case, we use $3\times 3$ depthwise convolution for its efficiency in parameters and FLOPs, where the $3\times 3$ convolutional filters are separated into each channel. 
Inspired by the data-dependent gating mechanism in GLA~\cite{gla}, we propose a data-dependent gating aggregation for 2D locality injection to interleave the global and local information:
\begin{align}
\begin{split}
    \mathbf{O}_{\text{local}} &= \mathrm{DWConv}_{3\times 3}(X),\\
    \mathbf{O}_{\text{global}} &= \mathrm{BiGLA}(\mathbf{O}_{\text{local}}),\\
    \mathbf{G}_{\text{2D}} &= \mathrm{sigmoid}(\mathbf{O}_{\text{local}} \bm{W}_{\text{gate2D}} + \bm{b}_{\text{gate2D}}),\\
    \mathbf{O} &= \mathbf{G}_{\text{2D}}\odot \mathbf{O}_{\text{local}}  + (\mathbf{1}-\mathbf{G}_{\text{2D}})  \odot \mathbf{O}_{\text{global}}.
\end{split}
\end{align}

\subsection{Architecture Details}

Equipped with the proposed \name{} block, we mainly investigate two kinds of variants of \name{}: ViT-style non-hierarchical models with a fixed number of tokens in each block and CNN-style hierarchical models with gradually downsampled tokens.

For ViT-style models, we set the patch size $p$ to 16 and stack 12 \name{} blocks. Then, we obtain 3 variants of the model at different sizes (\name{}-T, \name{}-S, and \name{}-B) by directly adjusting the embedding dimension $d$, which have similar parameters to DeiT-T, S, and B. For hierarchical models, we also propose 3 variants (\name{}-H-T, \name{}-H-S, and \name{}-H-B) following the design of SwinTransformer. We set the patch size $p$ to 4 and apply our proposed \name{} block at different stages.
The detailed architectures are provided in Appendix~\ref{app_sec:arch_details}.

\subsection{Efficient Implementation}
\label{sec:hardware}

The practical efficiency of the model not only depends on the theoretical FLOPs but is mostly determined by the hardware-awareness of the implementation. It means the implemented model should: 1) be aware of memory hierarchy; 2) leverage the specialized compute unit (tensor cores on GPU for fast matrix multiplication); 3) have a high degree of parallelism.
Thanks to the hardware-aware implementation of GLA, the most calculation-intensive parts of \name{} can be represented in matrix multiplication and are performed on faster SRAM instead of slower high bandwidth memory (HBM), leading to superior wall-time efficiency by leveraging the tensor cores and reducing the HBM I/O cost.

\boldparagraph{Hardware-aware Bidirectional Design.} Given the multi-directional nature of representing 2D images in 1D flattened sequences, the pioneering work Vim, based on bidirectional modeling, chooses to invoke two sequential kernels to process the forward and backward directions separately, which is inefficient in terms of parallelism. In this work, we propose a hardware-aware bidirectional design by fusing the forward and backward directions of Eq.~\eqref{eq:bigla} into a single kernel to achieve higher parallelism. Moreover, owing to the parameter-efficient design of the BiGLA layer, we can reduce the materialization of the backward visual sequence in high-bandwidth memory (HBM), which saves on memory costs.

\section{Experiment}
We conduct extensive experiments to validate the effectiveness of our proposed models. We present the main results on ImageNet~\cite{imagenet} and compare them with various other models. Additionally, we benchmark our model on downstream dense prediction tasks, including object detection on the COCO~\cite{coco} dataset and semantic segmentation on ADE20K~\cite{ade20k}.

\begin{table*}[htp!]
    \centering
    \caption{Comparison with plain non-hierarchical architectures (left) and hierarchical architectures (right) on ImageNet-1K validation set.
    ``Size'' means train/val image size. ``\#P.'', ``Tp.'', and ``Acc.'' denote the number of parameters, throughput and top-1 accuracy respectively. Tp. (images/s) is measured on a single 4090 GPU with batch size 256 following \cite{swin}.}
    \begin{minipage}[t]{0.48\textwidth}        
    \centering
    \label{tab:clscomp-nonhier}
    \vspace{-0.3cm}
    \small
    \renewcommand\tabcolsep{2pt}
    \resizebox{1.0\linewidth}{!}{
    \begin{tabular}{l| c  r r r| c }
    \toprule
    Method  &  Size & \#P. & FLOPs & Tp. & Acc. \\
    \toprule
    \multicolumn{6}{c}{\textbf{Transformers}} \\
    \midrule
    ViT-B/16~\cite{vit} &$384^{2}$ &  86M & 55.4G & -& 77.9  \\
    ViT-L/16~\cite{vit}  & $384^{2}$ & 307M  & 190.7G & -& 76.5 \\
    \midrule
    DeiT-T~\cite{deit} & $224^{2}$ & 6M  & 1.3G & 5761& 72.2  \\
    DeiT-S~\cite{deit} & $224^{2}$ & 22M  & 4.6G & 2396& 79.8 \\
    DeiT-B~\cite{deit} &  $224^{2}$ & 86M  & 17.6G & 837& 81.8 \\
    \toprule
    \multicolumn{6}{c}{\textbf{MLP}} \\
    \midrule
    gMLP-T~\cite{liu2021pay}& $224^{2}$ & 6M & 1.4G &  3872& 72.3 \\
    gMLP-S~\cite{liu2021pay} & $224^{2}$ & 20M & 4.5G & 1676& 79.6 \\
    gMLP-B~\cite{liu2021pay} & $224^{2}$ & 73M & 15.8G & 647& 81.6 \\
    \toprule
    \multicolumn{6}{c}{\textbf{SSMs}} \\
    \midrule
    S4ND-ViT-B~\cite{nguyen2022s4nd} & $224^{2}$ & 89M & - & 562& 80.4 \\
    \midrule
    Vim-T~\cite{vim} &$224^{2}$ & 7M & 1.5G & 2561& 76.1 \\ %
    Vim-S~\cite{vim} & $224^{2}$ & 26M & 5.1G & 1151& 80.3 \\
    \midrule
    LocalVim-T~\cite{localmamba} & $224^{2}$ & 8M & 1.5G & 885& 76.2 \\
    LocalVim-S~\cite{localmamba} & $224^{2}$ & 28M & 4.8G & 396& 81.2 \\
    \midrule
    PlainMamba-L1~\cite{yang2024plainmamba} & $224^{2}$ & 7M & 3.0G & 995& 77.9 \\
    PlainMamba-L2~\cite{yang2024plainmamba} & $224^{2}$ & 26M & 8.1G & 482& 81.6 \\
    PlainMamba-L3~\cite{yang2024plainmamba} & $224^{2}$ & 51M & 14.4G & 279& 82.3 \\
    \toprule
    \multicolumn{6}{c}{\textbf{Linear RNN}} \\
    \midrule
    VRWKV-T~\cite{vrwkv} &$224^{2}$&6M&1.2G&4551&75.1 \\
    VRWKV-S~\cite{vrwkv} &$224^{2}$&24M&4.6G&1724&80.1 \\
    VRWKV-B~\cite{vrwkv} &$224^{2}$&94M&18.2G&635&82.0 \\
    \toprule
    \multicolumn{6}{c}{\textbf{Linear Attention}} \\
    \midrule
    \rblue
    \name{}-T &$224^{2}$&6M&0.9G&4645&77.2 \\
    \rblue
    \name{}-S &$224^{2}$&23M&3.5G&1886&81.7 \\
    \rblue
    \name{}-B &$224^{2}$&89M&13.8G&701&82.6 \\
    
    \bottomrule
    \end{tabular}}
    \end{minipage}\hfill
    \begin{minipage}[t]{0.48\textwidth}
    \centering
    \small
    \label{tab:clscomp-hier}
    \vspace{-0.3cm}
    \renewcommand\tabcolsep{2pt}
    \resizebox{1.0\linewidth}{!}{
    \begin{tabular}{l | c  r r r| c }
    \toprule
    Method &  Size & \#P. & FLOPs & Tp. & Acc. \\
    \toprule
    \multicolumn{6}{c}{\textbf{Convnets}} \\
    \midrule
    RegNetY-4G~\cite{regnet} &  $224^{2}$ & 12M & 4G & - & 80.0\\
    RegNetY-8G~\cite{regnet} & $224^{2}$ & 25M & 8G & - & 81.7\\
    RegNetY-16G~\cite{regnet} & $224^{2}$& 45M & 16G & - & 82.9\\
    \midrule
    EffNet-B3~\cite{efficientnet} & $300^{2}$ & 12M & 1.8G & - & 81.6\\
    EffNet-B4~\cite{efficientnet} &$380^2$&19M&4.2G&-&82.9\\
    EffNet-B5~\cite{efficientnet} &$456^2$&30M&9.9G&-&83.6\\
    EffNet-B6~\cite{efficientnet} &$528^2$&43M&19.0G&-&84.0\\
    EffNet-B7~\cite{efficientnet} &$528^2$&66M&37.0G&-&84.3\\
    \midrule
    ConvNeXt-T~\cite{convnext} & $224^{2}$ & 29M & 4.5G &1505 & 82.1\\
    ConvNeXt-S~\cite{convnext} & $224^{2}$ & 50M & 8.7G &905 & 83.1\\
    ConvNeXt-B~\cite{convnext} & $224^{2}$ & 89M & 15.4G &643& 83.8\\
    \toprule
    \multicolumn{6}{c}{\textbf{Transformers}} \\
    \midrule
    Swin-T~\cite{swin} & $224^{2}$ & 28M & 4.6G &1511&81.3 \\
    Swin-S~\cite{swin} & $224^{2}$ & 50M & 8.7G &915&83.0 \\
    Swin-B~\cite{swin} & $224^{2}$ & 88M & 15.4G &661&83.5 \\
    \toprule
    \multicolumn{6}{c}{\textbf{SSMs}} \\
    \midrule
    S4ND-ConvNeXt-T~\cite{nguyen2022s4nd} & $224^{2}$ & 30M & - &643& 82.2 \\
    \midrule
    VMamba-T~\cite{vmamba}  & $224^{2}$ & 31M & 4.9G & 1161& 82.5 \\
    VMamba-S~\cite{vmamba}  & $224^{2}$ &50M& 8.7G & 779&83.6\\
    VMamba-B~\cite{vmamba} & $224^{2}$ &89M& 15.4G & 557&83.9\\
    \midrule
    EfficientVMamba-B~\cite{pei2024efficientvmamba} & $224^{2}$ & 33M & 4.0G & 1258& 81.8 \\
    \midrule
    LocalVMamba-T~\cite{localmamba} & $224^{2}$ & 26M & 5.7G & 330& 82.7 \\
    LocalVMamba-S~\cite{localmamba} & $224^{2}$ & 50M & 11.4G & 193 & 83.7 \\
    \toprule
    \multicolumn{6}{c}{\textbf{Linear Attention}} \\
    \midrule
    \rblue
    \name{}-H-T &$224^{2}$&29M&4.5G&1480&82.8 \\
    \rblue
    \name{}-H-S &$224^{2}$&50M&8.8G&890&83.8 \\
    \rblue
    \name{}-H-B &$224^{2}$&89M&15.5G&621&84.2\\
    
    \bottomrule
    \end{tabular}}
    
\end{minipage} 
\vspace{-0.3cm}
\end{table*}

\subsection{Image Classification}
\boldparagraph{Settings.} We train classification experiments on ImageNet-1K dataset, which is a widely used large-scale benchmark for image classification. To fairly compare with previous works, we mainly follow the training and evaluation setting of DeiT and SwinTransformer~\cite{deit,swin}. 
Specifically, all the models are trained from scratch for 300 epochs. Images are cropped to $224 \times 224$ for both training and evaluation. Further details are provided in Appendix~\ref{app_sec:exp_details}.

\boldparagraph{Comparison with Non-hierarchical Architectures.}
Tab.~\ref{tab:clscomp-nonhier} compares \name{} with plain non-hierarchical architectures based on different sequence modeling layers, including Transformer, SSM, and linear RNN. The results show that the proposed \name{} achieves superior trade-off in terms of parameters and accuracy across various model sizes, as shown in Fig.~\ref{fig:acc_comp} (a).
Remarkably, 
\name{}-S has nearly the same number of parameters as DeiT-S and significantly outperforms it by 1.9\% top-1 accuracy, which matches the performance of DeiT-B (only 0.1\% lower) with 3.7$\times$ fewer parameters, 5$\times$ fewer FLOPs and 2$\times$ faster throughput. Moreover, \name{}-B reaches 82.6\% top-1 accuracy, surpassing DeiT-B by 0.8\%, VRWKV-B by 0.6\%, and S4ND-ViT-B by 2.2\%.

In terms of practical throughput, \name{} surpasses other linear-complexity sequence modeling methods, notably being 1.8$\times$ faster than Vim-T and 1.6$\times$ faster than Vim-S at tiny and small model sizes respectively.
Compared with the counterparts with multi-direction scanning, \name{}-T is 5.2$\times$ faster than LocalVim-T and 4.6$\times$ faster than PlainMamba-L1 thanks to the memory-friendly access pattern of bidirectional modeling.
\name{} is only slightly slower than DeiT, which can be attributed to the high parallelism of the Transformer and the low FLOPs when inputting small $224\times 224$ images. As shown in  Fig.~\ref{fig:efficiency}, \name{} achieves exponential superiority as the image size increases.

\boldparagraph{Comparison with Hierarchical Architectures.}
Tab.~\ref{tab:clscomp-hier} compares \name{}-H with hierarchical architectures, including the advanced CNN-based RegNet and ConvNeXt, as well as Transformer-based Swin Transformer, and SSM-based S4ND and VMamba. Thanks to the linear complexity of the proposed VGLA block, \name{}-H achieves similar FLOPs to those of window-based Transformers and CNNs, but with the added advantage of a global receptive field. As shown in Fig.~\ref{fig:acc_comp} (b), \name{}-H achieves the best top-1 accuracy across different model sizes.

The results of practical throughput demonstrate that \name{}-H surpasses the SSM-based S4ND and VMamba, and matches the performance of well-established and highly-optimized ConvNeXt and SwinTransformer.

\subsection{Object Detection}
\boldparagraph{Settings.} We conduct experiments for object detection and instance segmentation on the COCO 2017 dataset~\cite{coco}. We utilize Mask-RCNN~\cite{he2017mask} as the detection head and follow VRWKV~\cite{vrwkv} to integrate the ViT-Adapter~\cite{chen2022vision} on our plain \name{} models. Further details are provided in Appendix~\ref{app_sec:exp_details}.

\boldparagraph{Results.} 
In Tab.~\ref{tab:detection}, for high-resolution $1333 \times 800$ input images, ViT needs to resort to window attention to ensure efficiency, sacrificing accuracy. Unlike ViT, \name{} can efficiently process high-resolution images directly with a global receptive field. The results demonstrate that \name{} outperforms both ViT and VRWKV in terms of FLOPs and accuracy. Specifically, \name{}-T uses only half the backbone FLOPs of ViT-T but achieves 1.7 higher $\rm AP^{b}$ and 1.2 higher $\rm AP^{m}$, surpassing VRWKV-T by 1.6 in  $\rm AP^{b}$ and 1.1 in $\rm AP^{m}$. For the base model size, though VRWKV-B is more efficient than ViT-B in FLOPs, it still falls short of ViT-B by 0.1 in $\rm AP^{m}$. Meanwhile, our \name{}-B achieves 0.5 higher $\rm AP^{b}$ and 0.4 higher $\rm AP^{m}$ than ViT-B with 44\% lower backbone FLOPs.

\begin{table*}[t!]
    \centering

  \caption{Object detection and instance segmentation on COCO val2017 (left) \& semantic segmentation on ADE20K val set (right).   ``\#Param'' denotes the number of backbone parameters. ``FLOPs'' in the left and right tables denote the computational workload of the backbone with an input image of $1333 \times 800$ and $512 \times 512$, respectively.
    ``$\dagger$'' means window attention is adopted in ViT layers.  ``$\ddagger$'' denotes that the results of Vim are directly taken from its paper since it uses the same segmentation head and training recipe as ours.
    }
    \vspace{-0.3cm}
    \begin{minipage}[t]{0.48\textwidth}
    \centering
    \label{tab:detection}

    \resizebox{.96\textwidth}{!}{
    \begin{tabular}{l|rrccc}
        \toprule
        Method       & \#Param.  & FLOPs & $\rm AP^{b}$ & $\rm AP^{m}$ \\
        \midrule
        ViT-T$^\dagger$ & 8M &   95.4G & 41.1 & 37.5  \\
        ViT-T          & 8M &  147.1G & 41.6 & 37.9  \\ 
        VRWKV-T & 8M &   67.9G & 41.7 & 38.0  \\
        \rblue
        \name{}-T & 8M &  \textbf{61.2G} & \textbf{43.3}	& \textbf{39.1}  \\
	\midrule
 
        ViT-S$^\dagger$ & 28M & 241.2G & 44.6 & 39.7  \\
        ViT-S           & 28M & 344.5G & 44.9 & 40.1  \\
        VRWKV-S  & 29M & 189.9G & 44.8 & 40.2  \\
        \rblue
        \name{}-S & 28M &  \textbf{164.1G} & \textbf{45.5} & \textbf{40.8}  \\
	\midrule
 
        ViT-B$^\dagger$ & 100M & 686.7G & 46.2 & 41.5  \\
        ViT-B           & 100M & 893.3G &  46.8   & 41.8  \\
        VRWKV-B & 107M& 599.0G & 46.8 & 41.7  \\
        \rblue
        \name{}-B & 103M &  \textbf{498.5G} & \textbf{47.3}& \textbf{42.2} \\
	\bottomrule
    \end{tabular}
    }
    \end{minipage}\hfill
    \begin{minipage}[t]{0.48\textwidth}
    \centering
    \label{tab:segmentation}
    \resizebox{.86\textwidth}{!}{

        \begin{tabular}{l|rrcc}
            \toprule
            Method                        & \#Param.  & FLOPs  & mIoU  \\
            \midrule
            \color{gray}Vim-T$^\ddagger$ & \color{gray}- &\color{gray}- & \color{gray}41.0\\
            ViT-T & 8M     & 20.9G  & 42.6  \\
            VRWKV-T & 8M     & 16.6G  & 43.3  \\
            \rblue
            \name{}-T & 8M & \textbf{14.9G} & \textbf{43.8}  \\
        \midrule
        \color{gray}Vim-S$^\ddagger$ & \color{gray}- &\color{gray}- & \color{gray}44.9 \\
            ViT-S & 28M    & 54.0G  & 46.2  \\
            VRWKV-S& 29M    & 46.3G  & 47.2  \\
            \rblue
            \name{}-S & 28M & \textbf{40.0G} & \textbf{47.9}  \\
        \midrule
     
            ViT-B & 100M    & 157.9G & 48.8  \\
            VRWKV-B & 107M   & 146.0G & 49.2  \\
            \rblue
            \name{}-B & 103M & \textbf{121.5G} & \textbf{49.4}  \\
        \bottomrule
        \end{tabular} 
    }
    \end{minipage} 
    \vspace{-0.4cm}
    \end{table*}

\subsection{Semantic Segmentation}
\boldparagraph{Settings.} We train experiments for semantic segmentation on the ADE20K~\cite{ade20k} dataset. We use UperNet~\cite{xiao2018unified} as the segmentation head and adopt the ViT-Adapter~\cite{chen2022vision} following VRWKV~\cite{vrwkv} to adapt our plain \name{} for segmentation. Training details are the same as VRWKV~\cite{vrwkv} and Vim~\cite{vim}, which are detailed in the Appendix~\ref{app_sec:exp_details}.

\boldparagraph{Results.} As shown in Tab.~\ref{tab:segmentation}, for medium-resolution $512 \times 512$ input images, \name{} outperforms the quadratic-complexity Transformer-based ViT and the linear-complexity RNN-based VRWKV across different model sizes in both FLOPs and segmentation accuracy. For instance, in the tiny size models, \name{} outperforms ViT by 1.2 mIoU and 5G FLOPs, and VRWKV by 0.5 mIoU and 1.7G FLOPs. In the small size models, \name{} surpasses ViT by 1.7 mIoU and 14G FLOPs, and VRWKV by 0.7 mIoU and 6.3G FLOPs. These results demonstrate \name{}'s superior adaptability for dense-pixel prediction tasks compared to VRWKV.

\begin{figure}[ht!]

    \centering
     \begin{subfigure}{0.48\textwidth}
         \centering
         \includegraphics[width=\textwidth]{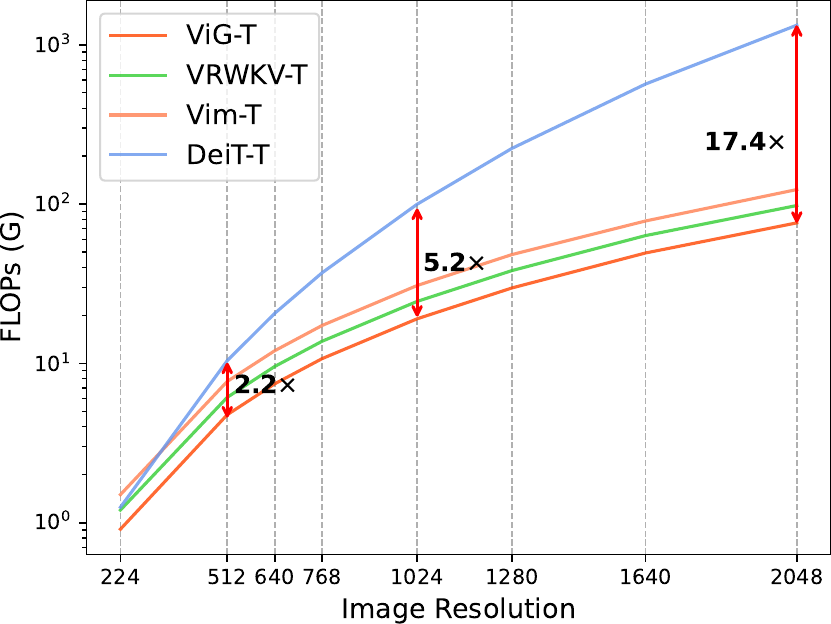}
         \vspace{-0.5 cm}
         \caption{FLOPs}
     \end{subfigure}
     \hfill
     \begin{subfigure}{0.475\textwidth}
         \centering
         \includegraphics[width=\textwidth]{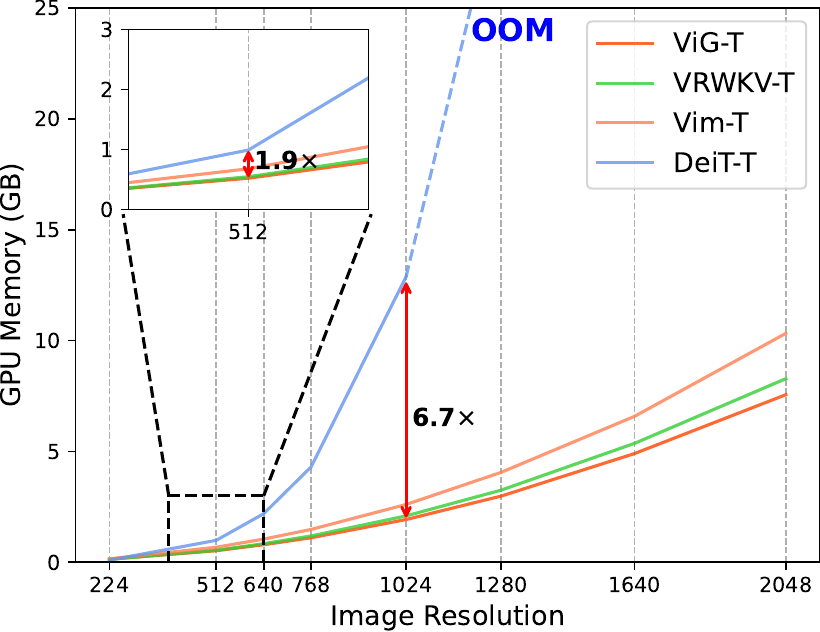}
         \vspace{-0.5 cm}
         \caption{Memory}

        \end{subfigure}

     \begin{subfigure}{0.48\textwidth}
         \centering
         \includegraphics[width=\textwidth]{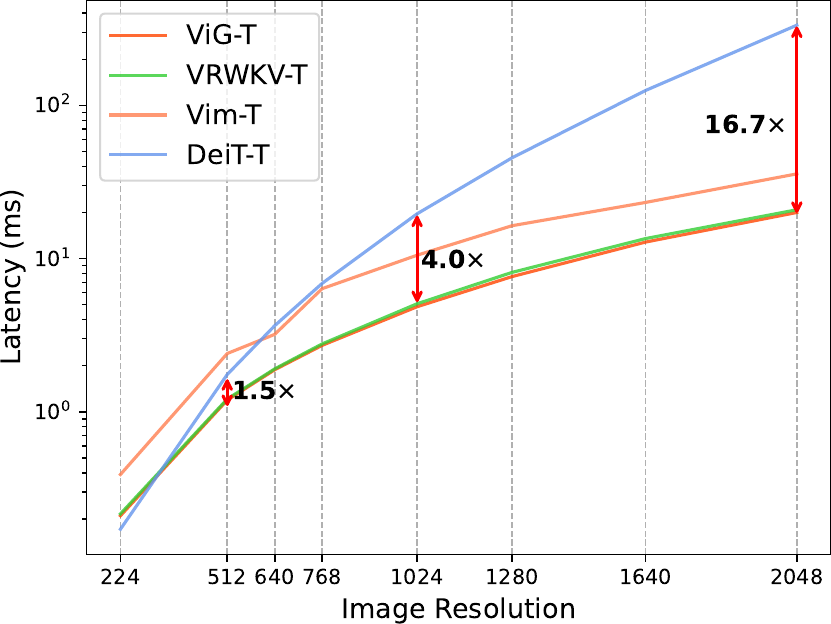}
         \vspace{-0.5 cm}
         \caption{Latency}
         
        \end{subfigure}
     \hfill
     \begin{subfigure}{0.475\textwidth}
         \centering
         \includegraphics[width=\textwidth]{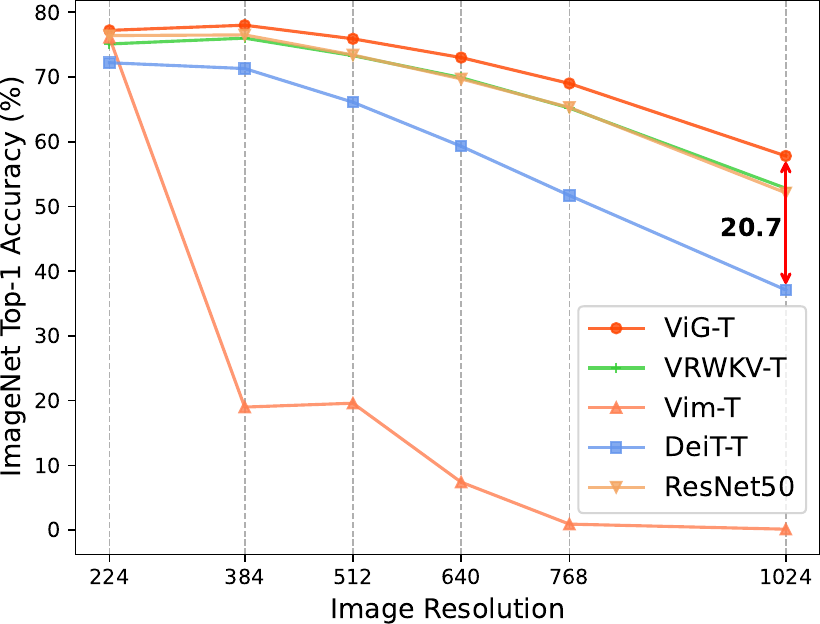}
         \vspace{-0.5 cm}
         \caption{Accuracy}
        \end{subfigure}
        \vspace{-0.2cm}
        \caption{Comparison among \name{}, Vim~\cite{vim}, VRWKV~\cite{vrwkv}, and ViT~\cite{vit,deit} in (a) FLOPs, (b) memory, (c) latency, and (d) accuracy with respect to increasing image resolution during inference on ImageNet-1K val set. The blue dashed line indicates the estimated values when the GPU memory has run out. We benchmark the latency with the maximum batch size that can make models runnable on the GPU to ensure full GPU utilization and provide available results at high resolutions.} 
        \label{fig:efficiency}
    
    \vspace{-0.6cm}
\end{figure}

\begin{table}[t!]
    \vspace{-0.2cm}
    \centering
    \caption{\textbf{Roadmap of \name{}.} The throughput and memory are measured on a single NVIDIA RTX 4090 GPU with batch size 256 and image size 224 following \cite{swin}. }
    \label{tab:roadmap}

    \resizebox{.92\textwidth}{!}{

    \begin{tabular}{ll|rrrc}
    \toprule
    Method & Roadmap & \#Param. & Throughput & Memory & Top-1 Acc.\\
    \midrule
    DeiT & - & 5.72M & 5761 & 627MB & 72.2 \\
    \midrule
    GLA & - & 5.72M & 6662 & 582MB & 66.1 \\
     $+$ & improved patch embedding layer  & 5.75M & 6634 & 582MB & 73.8 \\
     $+$ & direct bidirectional modeling  & 5.75M & 4564 & 748MB & 75.2 \\
     $+$ & absolute position embedding  & 5.79M & 4571 & 748MB &  75.4 \\
     $+$ & direction-wise $\overline{\alpha}$ (\cref{sec:bigla}) &5.81M & 4566 & 785MB& 76.3 \\
     $+$ & 2D gating locality injection (\cref{sec:2dgating})  & 5.83M & 3812 & 842MB& 77.2 \\
     $+$ & hardware-aware bidirection impl. (\cref{sec:hardware}) & 5.83M & 4645 & 730MB & 77.2 \\
    \bottomrule
    \end{tabular} 
    }
    \vspace{-0.3cm}
    \end{table}
\subsection{Ablation Study}

More ablation studies are provided in Appendix~\ref{app_sec:ablation}.

\boldparagraph{Roadmap.}
In Tab.~\ref{tab:roadmap}, we show the roadmap of how to introduce minimal cost to render the causal sequence modeling GLA into proposed \name{}. 
The ``improved patch embedding layer'' of Row 3 means that instead of directly applying convolution with large $16\times16$ kernel, we adopt a $9\times9$ convolution with stride as 8 followed by $3\times3$ convolution with stride as 2, which only adds 0.03  M parameters but boost the accuracy with nearly no affects on inference efficiency.
The \name-T introduces only 0.11M parameters and significantly outperforms vanilla GLA by 11.1\% top-1 accuracy and DeiT by 5.0\% top-1 accuracy.
By further introducing the hardware-aware bidirectional implementation, we significantly boost the efficiency (enhance throughput by 21.9\% and save 13.3\% GPU memory) and close the gap with DeiT, even at the low-resolution $224 \times 224$ image.

\boldparagraph{Efficiency of \name{}.}
In Fig.~\ref{fig:efficiency}, we compare \name{}-T with Vim-T, VRWKV-T, and DeiT-T, focusing on theoretical FLOPs, actual latency, and memory usage on an RTX 4090 GPU. We test the complete model across increasing input image resolutions from $224 \times 224$ to $2048 \times 2048$. Thanks to the linear-complexity sequence modeling, the advantage of the proposed architecture over ViT grows as the resolution increases. 
When resolution reaches $1024 \times 1024$, \name{}-T uses 5.2$\times$ lower FLOPs, saves 90\% GPU memory, and runs 4.8$\times$ faster than DeiT-T.
Compared to its linear-complexity counterpart Vim, \name{} also demonstrates 1.6$\times$ lower FLOPs, saves 26\% GPU memory, and runs 2.2$\times$ faster.
Additionally, \name{} surpasses VRWKV in terms of FLOPs, latency, and memory.

\boldparagraph{Accuracy  \vs Resolution.} 
In Fig.~\ref{fig:efficiency} (d), we test the models trained on $224 \times 224$ resolution across different resolutions. \name{} outperforms ViT, Vim, VRWKV, and even hierarchical CNN-based ResNet50 in terms of accuracy as the resolution increases. The results demonstrate that \name{} benefits from better 2D-awareness and demonstrates superior generalization in resolution extrapolation.

\section{Conclusion}
In this paper, we introduced \name{}, a generic vision backbone network that introduces Gated Linear Attention (GLA) to the vision field, achieving efficient visual representation learning. Our approach addresses the inherent limitations of traditional Transformers and CNNs by maintaining a global receptive field while operating with linear complexity. 
We propose direction-wise gating through bidirectional GLA modeling and 2D gating locality injection to effectively capture both global context and local details, leading to significant improvements in accuracy. Additionally, our hardware-aware implementation reduces the overhead brought by extra direction,  enhancing efficiency.
The superior experimental results of \name{} in low-resolution image classification, medium-resolution segmentation, and high-resolution detection highlight it as a very competitive alternative to the existing generic vision backbones.

\boldparagraph{Limitations.}
Though \name{} proposes hardware-aware implementation improves the efficiency, as shown in Tab.~\ref{tab:roadmap}, and demonstrates obvious superiority at high-resolution input images, it is still slightly inferior to DeiT at the small $224\times 224$ input images. We will further optimize the implementation to improve the hardware-awareness in future work.

\section*{Acknowledgement}
We would like to acknowledge Yuxin Fang for helpful feedback on the draft.

{
\small
\bibliographystyle{plain}
\bibliography{refbib}
}

\clearpage
\appendix

\section{Extended Related Work}
\label{app_sec:related}
\boldparagraph{Visual Representation Learning through CNN and Transformer.}
Visual representation learning remains a cornerstone of computer vision research, substantially propelled by advancements in deep neural networks. Convolutional Neural Network (CNN)~\cite{cnn} has long been the dominant architecture in this field for its efficient processing of spatial data. It serves as the backbone in many advanced vision encoders~\cite{resnet,resnext,convnext,efficientnet,regnet,densenet,hrnet,xie2017aggregated,krizhevsky2012imagenet,redmon2018yolov3,ding2022scaling,ding2023unireplknet,liu2022more,hou2022conv2former,wang2023internimage,rao2022hornet,howard2017mobilenets,ma2018shufflenet,tan2021efficientnetv2} across various tasks. Vision Transformer (ViT)~\cite{vit}, a seminal work, challenges the dominance of CNNs by introducing the plain, non-hierarchical Transformer architecture~\cite{vaswani2017attention} to the vision domain, demonstrating that visual representation learning can effectively be conducted in a pure sequence-to-sequence manner. DeiT~\cite{deit} further refines ViT by enhancing optimization through advanced training techniques and distillation. To mitigate the notorious issue of quadratic complexity in Transformer softmax attention, the SwinTransformer~\cite{swin} introduces a hierarchical structure and confines attention computation to local windows, effectively reducing complexity to linear. Similarly, Pyramid Vision Transformer (PVT)~\cite{pvt} employs spatial downsampling on key and value feature maps to decrease the computational demands of global attention. Numerous subsequent studies~\cite{wu2021cvt,wang2022pvt,yang2021focal,dai2021coatnet,d2021convit,fang2022msg,guo2022cmt,dong2022cswin,tu2022maxvit,zhang2022topformer,liu2022swin,yuan2021tokens,chu2021twins,zhang2022hivit,li2023uniformer} continue to enhance ViT performance, drawing inspiration from the successes of CNNs.

\boldparagraph{Linear Complexity Sequence Modeling.}
Transformer~\cite{vaswani2017attention} has been very successful in sequence modeling, serving as a core module in many state-of-the-art Large Language Models (LLMs)~\cite{touvron2023llama,brown2020language,radford2019language,bai2023qwen} in the NLP field. However, it requires retaining all the historical Key-Value cache and attending the current token to all the historical caches, limiting its application for long sequences. 
To circumvent this issue, RNN-like sequence modelings~\cite{hochreiter1997long,qin2024hierarchically,linearattn,choromanski2020rethinking,arora2024simple,qin2024hgrn2,dai2019transformer,de2024griffin,munkhdalai2024leave,ma2024megalodon} attract increasing interest for its merits in linear complexity, fixed hidden state, and global context interaction. 
RWKV~\cite{rwkv} and RetNet~\cite{retnet} introduce temporal decay to model the long-range dependency. Recently, building on SSM~\cite{gu2021efficiently,smith2023simplified,fu2023hungry,mehta2023long}, Mamba~\cite{mamba} introduces a novel selective SSM operation and demonstrates competitive performance against modern optimized Transformer models with linear complexity.
To further improve the practical efficiency, Mamba introduces hardware-aware implementation to reduce the I/O and memory costs. Drawing inspiration from linear attention~\cite{linearattn,mao2022fine,qin2022devil} and retention~\cite{retnet}, GLA~\cite{gla} adds a novel data-dependent gating mechanism to enhance the expressiveness and implements it in a hardware-aware manner~\cite{dao2022flashattention,dao2023flashattention} to significantly boost efficiency.

\boldparagraph{Linear Complexity Visual Sequence Modeling.} 
Inspired by the success of ViT~\cite{vit} in visual sequence learning and Mamba~\cite{mamba} in achieving linear complexity in language modeling, Vim~\cite{vim} introduces the advanced Mamba block to vision by replacing the quadratic softmax attention in ViT with the proposed linear complexity bidirectional SSM~\cite{schuster1997bidirectional,wang2022pretraining,yan2023diffusion}. This adaptation demonstrates competitive performance with improved efficiency compared to ViT. Concurrently, VMamba~\cite{vmamba} adopts a hierarchical macro architecture similar to SwinTransformer~\cite{swin} and proposes using the Mamba block to scan the 1D visual sequence in criss-cross patterns~\cite{huang2019ccnet,ho2019axial} for 2D visual representation learning. Numerous follow-up works~\cite{zigma,yang2024plainmamba,localmamba,pei2024efficientvmamba,li2024mamba} explore the use of Mamba to scan the 1D visual sequence in more complex 2D-aware patterns. For instance, PlainMamba~\cite{yang2024plainmamba} proposes multiple continuous 2D scannings for 1D visual sequence, while LocalMamba~\cite{localmamba} opts for scanning within local windows. In contrast, VisionRWKV~\cite{vrwkv} forgoes the Mamba block, adapting the linear complexity RWKV block from NLP for use in vision.

\section{Architecture Details}
\label{app_sec:arch_details}
We provide detailed configuration of our non-hierarchical \name{} and hierarchical \name{}-H in Tab.~\ref{tab:arch}.

\begin{table}[h!]
    \centering
  \caption{\textbf{Detailed configurations of different variants of \name{}.} For hierarchical variants, we provide the number of channels and blocks in 4 stages. The FLOPs are calculated with $224\times 224$ input image.} \vspace{5pt}
    \begin{tabular}{l c  c c r r}
    \toprule
      Model & \#Blocks &  \#Channels &  \#Heads & Params & FLOPs  \\ \midrule
      \name{}-T &  12 & 192  & 3 & 6M & 0.9G\\
      \name{}-S &  12 & 384  & 6 & 23M& 3.5G\\
      \name{}-B &  12 & 768  & 12 & 89M& 13.8G\\
        \midrule
      \name{}-H-T &  [2, 2, 5, 2] & [96, 192, 384, 768] & [ 3, 6, 12, 24 ] & 29M & 4.5G\\
      \name{}-H-S &  [2, 2, 17, 2] & [96, 192, 384, 768] &[ 3, 6, 12, 24 ] & 50M & 8.8G\\
      \name{}-H-B &  [2, 2, 17, 2] & [128, 256, 512, 1024] &[4, 8, 16, 32] & 89M & 15.5G\\
        \bottomrule
      \end{tabular}%
    \label{tab:arch} 
  \end{table}%

\section{Experimental Details}
\label{app_sec:exp_details}
\boldparagraph{ImageNet Classification Experimental Details.}
ImageNet contains 1.2M training images and 50K validation images from 1000 categories. We train the models on the training set and report the  top-1 accuracy on the validation set. 
All the model are trained from scratch for 300 epochs, with a cosine schedule and EMA, using a total batch size of 1024. We use AdamW optimizer~\cite{kingma2014adam} and set betas to (0.9, 0.999),  momentum to 0.9, weight decay to 0.05, initial learning rate to $1 \times 10^{-3}$. Images are cropped to $224 \times 224$ for both training and evaluation.
We build upon PyTorch~\cite{paszke2019pytorch} and train Tiny and Small models with 8$\times$ 4090 GPUs, and the Base model with 16$\times$ 4090 GPUs.

\boldparagraph{ADE20K Semantic Segmentation Experimental Details.}
All the models are initialized with ImageNet-1K pre-trained weights. Training details follow the setting in VRWKV~\cite{vrwkv}. We employ the AdamW optimizer, setting initial learning rate to $6\times10^{-5}$ for the Small/Base models and $1.2\times10^{-4}$ for the Tiny model, a batch size of 16, and a weight decay of 0.01. We train all the models for 160k iterations on the training set of the ADE20K dataset.

\boldparagraph{COCO Object Detection Experimental Details.}
All models are initialized with ImageNet-1K pre-trained weights and trained according to a 12-epoch schedule with a batch size of 16. We follow VRWKV~\cite{vrwkv} and employ the AdamW optimizer with a learning rate of $1 \times 10^{-4}$ and a weight decay of 0.05.

\section{Ablation Study}
\label{app_sec:ablation}

\begin{table}[t!]
  \centering
  \caption{\textbf{Comparison of different bidirectional design.} GLA$_{\text{vim}}$ denotes that we follow Vim to build bidirectional modeling on GLA by introducing an extra backward GLA layer. The throughput and memory are measured on a single RTX 4090 GPU with batch size 256 and image size 224 following \cite{swin}.}
    \begin{tabular}{l|rcrc}
      \toprule
      Method & \#Param. & Throughput & Mem. & Top-1 Acc.\\
      \midrule
      GLA & 5.72M & 6662 & 582MB &66.1 \\
      GLA$_{\text{Vim}}$ & 6.70M & 4524 & 812MB &73.5\\
      \rblue
      \name{} & 5.83M & 4645 & 730MB & 77.2 \\
      \bottomrule
      \end{tabular} 
  \label{tab:bid_comp}
\end{table}%
      
\boldparagraph{Bidirectional Design.}
As shown in Tab.~\ref{tab:roadmap}, the proposed bidirectional design by applying direction-wise gating $\overline{\alpha}$ introduces only 0.02M parameters but achieves an improvement of 0.9\% in top-1 accuracy without compromising efficiency. 
In Tab.~\ref{tab:bid_comp}, we introduce another variant, GLA$_{\text{Vim}}$, which follows the bidirectional design of Vim~\cite{vim} by incorporating an extra backward GLA layer, adding nearly 1M parameters. The results demonstrate that the proposed bidirectional modeling and implementation is faster, more parameter-efficient, more memory-efficient, and more accurate than Vim.

\begin{figure}[h!]
    \begin{center}
        \includegraphics[width=.98\textwidth]{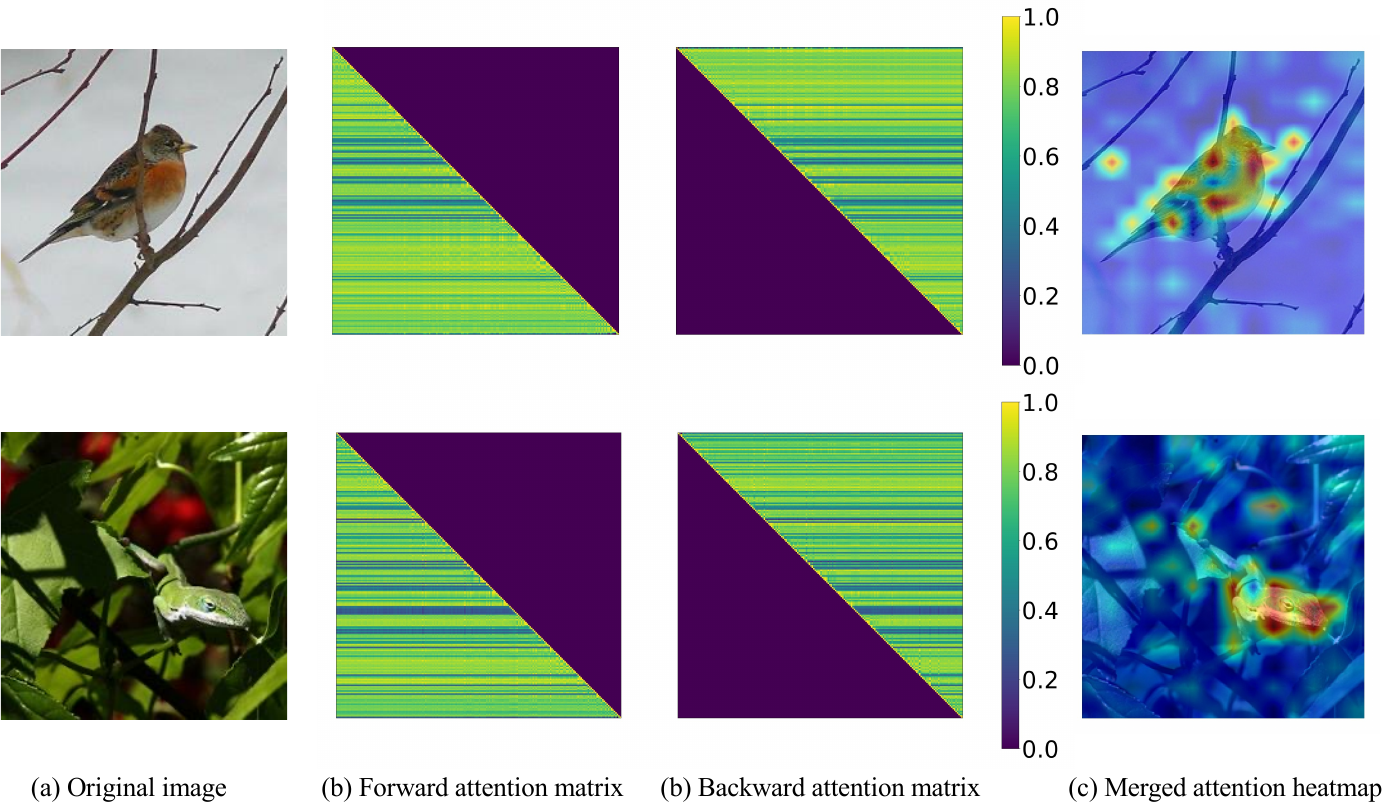}
        \caption{Visualization of attention maps.} 
        \label{fig:attn}
    \end{center}
\end{figure}

\boldparagraph{Gating Mechanism Understanding.}
In Fig.~\ref{fig:attn}, we visualize the attention maps of the proposed BiGLA layer. The results show that the forward and backward attention complement each other, and the merged attention can capture the prominent 2D context of the images.

\begin{figure}[t!]
    \begin{center}
        \includegraphics[width=.98\textwidth]{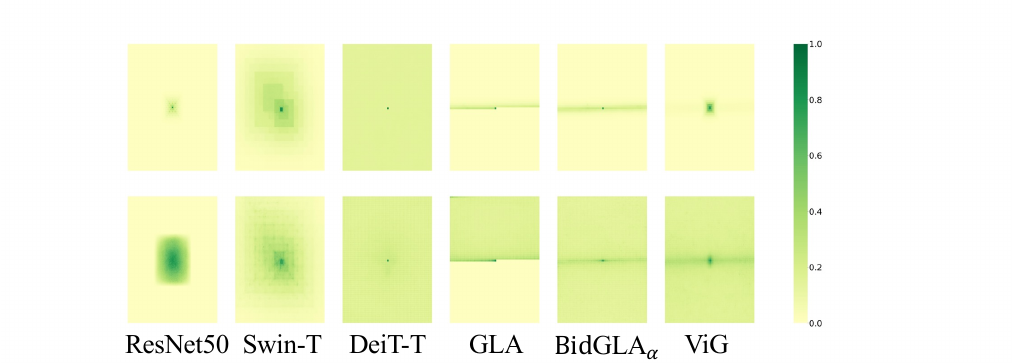}
        \caption{Comparison of Effective Receptive Field (ERF). ``BidGLA$_{\overline{\alpha}}$'' denotes the proposed bidirectional design with direction-wise gating $\overline{\alpha}$ (Row 4 in Tab.~\ref{tab:roadmap}). The proposed \name{} and BidGLA$_{\overline{\alpha}}$ exhibit a global ERF like DeiT. By introducing the 2D locality injection, the area of intensive response is larger.} 
        \label{fig:erf}
    \end{center}
\end{figure}

\boldparagraph{Effective Receptive Field Analysis.}
In Fig.~\ref{fig:erf}, we compare the effective receptive field (ERF) of different models, including CNN-based ResNet50, window-attention-based SwinTransformer, vanilla ViT variant DeiT. The results show that the \name{} exhibits a global ERF like DeiT, while the CNN-based and window-attention-based methods fail. We also visualize the ERF of variant BidGLA$_{\overline{\alpha}}$, which only applies direction-wise gating $\overline{\alpha}$ (Row 6 in Tab.~\ref{tab:roadmap}). The results demonstrate that the proposed 2D locality injection enlarges the area of intensive response, which is beneficial for capturing the spatial-aware context.

\end{document}